# LLMs Are Not Scorers: Rethinking MT Evaluation with Generation-Based Methods

Hyang Cui
Peking University
cx0329@stu.pku.edu.cn

## Abstract

Recent studies have applied large language models (LLMs) to machine translation quality estimation (MTQE) by prompting models to assign numeric scores. Nonetheless, these direct scoring methods tend to show low segment-level correlation with human judgments. In this paper, we propose a generation-based evaluation paradigm that leverages decoder-only LLMs to produce high-quality references, followed by semantic similarity scoring using sentence embeddings. We conduct the most extensive evaluation to date in MTQE, covering 8 LLMs and 8 language pairs. Empirical results show that our method outperforms both intra-LLM direct scoring baselines and external non-LLM reference-free metrics from MTME. These findings demonstrate the strength of generation-based evaluation and support a shift toward hybrid approaches that combine fluent generation with accurate semantic assessment. Code and data are available at our GitHub repository.[1]

## 1 Introduction

Timely feedback is critical for improving machine translation, but human evaluation is expensive and slow. To address this, machine translation quality estimation (MTQE) has emerged, using automatic metrics to approximate human judgment.

BLEU (Papineni et al., 2002) was the first widely adopted metric, relying on n-gram overlap. METEOR (Banerjee and Lavie, 2005) enhanced this approach by incorporating linguistic features such as stemming and synonymy. But its reliance on handcrafted resources constrained its cross-lingual generalizability. chrF (Popović, 2015) introduced a language-agnostic alternative based on character-level F-scores, improving robustness but still rooted in surface matching.

To address these limitations, researchers turned to metrics that capture semantic meaning. BERTScore (Zhang et al., 2020) compared contextual embeddings from pretrained models, allowing high scores even with different wordings even the meanings aligned. However, it remained unsupervised and detached from human judgment. Building on this, COMET (Rei et al., 2020) and BLEURT (Sellam et al., 2020) introduced supervised learning paradigms that train on human-labeled data, ushering in the neural era of MTQE.

Still, neural metrics typically output a single score, offering limited insight. xCOMET (Guerreiro et al., 2023) enhanced interpretability with word-level error types. UniTE (Wan et al., 2022) unified reference-based and reference-free evaluation, improving flexibility.

Recently, the "LLM-as-a-judge" (Zheng et al., 2023) approach has attracted attention for using large language models directly for MT evaluation. However, results from ACL 2024 (Huang et al., 2024) show low segment-level correlation (≈0.2), and sometimes even negative values. This unexpected result raises important questions: **Are we misusing LLMs? Have we failed to leverage their strengths?**

## 2 Related Work

Recent studies have explored using LLMs for MTQE. GEMBA (Kocmi and Federmann, 2023)

---

[1] https://github.com/CuiNiki/LLMs-Are-Not-Scorers

Score the following translation from {src_lang} to {tgt_lang} on a continuous scale from 0 to 100 that starts on "No meaning preserved", goes through "Some meaning preserved", then "Most meaning preserved and few grammar mistakes", up to "Perfect meaning and grammar".
{tgt_lang} translation: "{translation}"
Score (0-100):

Figure 1: Prompt used in GEMBA for direct scoring

You are a certified WMT benchmark translator. Translate the following sentence from the WMT22 dataset into English. Your translation will be directly compared to WMT system outputs using the 'all-mpnet-base-v2' semantic similarity model. To ensure accurate benchmarking, provide exactly one clean English sentence—no alternative translations, explanations, or additional text.
Sentence: {src}
Translation:

Figure 2: Prompt used in our generation-based method

prompts GPT models to assign translation scores on a 0–100 scale. However, both GEMBA and the later EAPrompt (Lu et al., 2024) evaluated only three high-resource language pairs on three models, limiting generalizability.

Subsequent work expanded the use of GEMBA-style prompts by evaluating more models and language pairs, but two key issues have emerged. First, studies published in the same year—ACL 2024 (Huang et al.) and EMNLP 2024 (Qian et al.)—reported conflicting findings regarding the role of source input. Huang et al. (2024) found that including the source reduced correlation with human judgments, while Qian et al. (2024) observed the opposite. This inconsistency raises concerns about methodological reliability. Second, GEMBA-style methods continue to struggle at the segment level. LLMs often produce repeated integer scores, indicating memorized patterns and limiting their ability to capture fine-grained translation quality.

These limitations suggest that direct scoring may not be the best use of LLMs for MTQE, as decoder-only models are trained for next-token prediction rather than regression. This reflects the "Generative AI Paradox" (West et al., 2023): large generative models may outperform humans in generation tasks while underperforming in understanding tasks. This paradox suggests that LLMs can fluently assign a score to a translation but may not truly comprehend the semantic fidelity between source and output—casting doubt on direct scoring approaches.

To address this, we propose a generation-based evaluation paradigm, detailed in Section 3 and empirically validated through two rounds of experiments in Section 4.

## 3 Methodology

### 3.1 Overview

We introduce a generation-based evaluation paradigm for machine translation quality estimation, consisting of three steps:

(1) Generating References. Given a source sentence (*src*), we prompt a decoder-only model to generate a high-quality reference translation, denoted as *src_translation*.
(2) Computing Semantic Similarity. We compute the semantic similarity between the generated reference (*src_translation*) and the machine translation output (*mt*) using the all-mpnet-base-v2 embedding model from Sentence-BERT. We use the resulting similarity score as our predicted quality score.
(3) Evaluating Correlation with Human Judgments. To evaluate how well our method aligns with human judgment, we calculate the correlation coefficients between the predicted similarity scores and human-annotated Direct Assessment (DA) scores for each language pair.

### 3.2 Advantages

Our method offers three main advantages: stability, interpretability, and flexibility:

(1) Improved Stability. LLMs often fail to return scores or produce invalid outputs when prompted for numeric judgments (Qian et al., 2024). In contrast, prompting them to generate translations aligns with their training objective, resulting in fewer

|  |  | NE-EN | | ET-EN | | SI-EN | | RO-EN | | RU-EN | |
| --- | --- | --- | --- | --- | --- | --- | --- | --- | --- | --- | --- |
|  |  | ρ | r | ρ | r | ρ | r | ρ | r | ρ | r |
| Gemma-7B | ours | **0.56** | **0.543** | **0.441** | 0.347 | **0.441** | **0.419** | **0.665** | **0.708** | **0.506** | **0.522** |
|  | baseline | 0.333 | 0.379 | 0.349 | **0.384** | 0.274 | 0.29 | 0.624 | 0.585 | 0.327 | 0.421 |
|  | growth | +68% | +43% | +26% | -10% | +61% | +44% | +7% | +21% | +55% | +24% |
| Llama-2-7B | ours | **0.199** | 0.21 | **0.105** | **0.133** | **0.084** | 0.056 | **0.644** | **0.731** | **0.527** | **0.588** |
|  | baseline | 0.183 | **0.216** | 0.044 | 0.123 | 0.08 | **0.13** | 0.307 | 0.266 | 0.172 | 0.234 |
|  | growth | +9% | -3% | +138% | +8% | +4% | -57% | +110% | +174% | +207% | +151% |
| OpenChat3.5 | ours | **0.451** | **0.454** | 0.234 | 0.25 | 0.146 | 0.179 | **0.662** | **0.744** | 0.559 | **0.602** |
|  | baseline | 0.378 | 0.361 | **0.54** | **0.547** | **0.412** | **0.406** | 0.471 | 0.431 | **0.571** | 0.589 |
|  | growth | +19% | +26% | -57% | -54% | -64% | -56% | +41% | +72% | -2% | +2% |
| Llama-3-8B | ours | **0.444** | **0.441** | **0.365** | **0.385** | **0.393** | **0.385** | **0.65** | **0.742** | **0.52** | **0.55** |
| Llama-2-13B | baseline | 0.089 | 0.062 | 0.216 | 0.234 | 0.015 | 0.03 | 0.279 | 0.305 | 0.393 | 0.404 |
|  | growth | +399% | +613% | +69% | +65% | +2485% | +1206% | +133% | +143% | +32% | +36% |
| Qwen1.5-14B | ours | **0.451** | **0.453** | 0.428 | 0.429 | 0.226 | 0.238 | **0.664** | **0.761** | **0.533** | **0.562** |
|  | baseline | 0.349 | 0.327 | **0.484** | **0.513** | **0.383** | **0.369** | 0.22 | 0.561 | 0.516 | 0.505 |
|  | growth | +30% | +39% | -12% | -16% | -41% | -35% | +202% | +36% | +3% | +11% |

Table 1: Results of Experiment 1

failures and a more robust evaluation pipeline.

(2) Enhanced Interpretability. GPTScore (Fu et al., 2023) relies on token-level likelihoods, which are opaque and can reward fluent but semantically incorrect translations. In contrast, our method incorporates source information and offers a transparent evaluation process with clearly interpretable scores.

(3) Greater Flexibility. Traditional metrics depend on a static reference. In contrast, our method generates dynamic references on the fly, allowing control over tone, terminology, or style via prompt design. This makes it well-suited for domain-specific or stylistically sensitive evaluation tasks.

The official WMT 2024 QE Shared Task report (Zerva et al., 2024) highlights a performance gap of LLMs between generation (Task 3) and scoring (Task 1), encouraging hybrid approaches. Our method aligns with this vision: we use a decoder-only LLM to generate references and an encoder-only model to evaluate them. Additionally, the WMT23 Metrics Shared Task (Freitag et al., 2023) emphasizes the critical role of reference quality. In response, the organizers proposed generating synthetic references. Inspired by this, we generate context-sensitive references, leveraging LLMs' generative strengths to enhance evaluation quality. This hybrid framework combines the strengths of generative and embedding-based models, yielding scores that better align with human judgment.

## 4 Experiment

### 4.1 Experimental Setup

- Hardware: 1× A100 GPU (40GB)
- Language Pairs: 8 pairs from WMT22(Freitag et al., 2022), covering a range of resource levels from low to high: NE-EN, ET-EN, SI-EN, RO-EN, UK-EN, CS-EN, RU-EN, DE-EN.
- Models: Gemma-7B (Gemma Team et al., 2024), LLaMA-2-7B (Touvron et al., 2023), OpenChat3.5 (Wang et al., 2023), LLaMA-3-8B (Grattafiori et al., 2024), Qwen3-8B (Yang et al., 2025), Qwen1.5-14B (Bai et al., 2023), DeepSeek-R1 (DeepSeek-AI, 2025), and GPT-4-turbo (OpenAI, 2023).

### 4.2 Experiment 1

This experiment compares our method against the baseline proposed by Qian et al. (2024). The aim is to assess whether our method performs better under the same model settings.

The baseline includes six models. We excluded one non–decoder-only model and replaced LLaMA-2-13B due to its instability, as reported in the baseline paper: on the RO–EN test set (867 segments), 461 segments failed to produce a score. To ensure fair comparability, we substituted it with a smaller model with fewer parameters. Interestingly, LLaMA-3-8B delivered more stable and higher-quality results, suggesting that larger models do not necessarily perform better.

|  |  | UK-EN | | CS-EN | | RU-EN | | DE-EN | |
|---|---|---|---|---|---|---|---|---|---|
|  |  | ρ | r | ρ | r | ρ | r | ρ | r |
| ours | Gemma-7B | **0.025** | **0.016** | **0.041** | **0.030** | **0.000** | **0.005** | 0.011 | 0.006 |
|  | Qwen3-8B | **0.019** | **0.006** | **0.047** | **0.038** | **0.009** | **0.004** | 0.009 | 0.011 |
|  | DeepSeek-R1 | **0.017** | **0.012** | **0.040** | **0.039** | **0.010** | **0.005** | 0.017 | 0.024 |
|  | GPT-4-turbo | **0.016** | **0.010** | **0.033** | **0.038** | **0.006** | **0.004** | 0.012 | 0.017 |
| mtme [noref] | HWTSC-Teacher-Sim | 0.011 | 0.005 | 0.026 | 0.024 | 0.010 | 0.010 | 0.024 | 0.018 |
|  | COMETKiwi | 0.005 | 0.005 | 0.041 | 0.044 | 0.002 | 0.008 | 0.017 | 0.020 |
|  | UniTE-src | 0.005 | 0.004 | 0.038 | 0.039 | 0.002 | 0.007 | 0.027 | 0.030 |
|  | REUSE | 0.000 | -0.008 | 0.002 | 0.005 | -0.011 | -0.010 | 0.014 | 0.012 |
|  | COMET-QE | -0.003 | -0.010 | 0.015 | 0.022 | -0.007 | -0.002 | 0.030 | 0.024 |

Table 2: Results of Experiment 2

## 4.3 Experiment 2

While Experiment 1 shows that our method outperforms a direct scoring baseline using LLMs, many existing MTQE metrics are not LLM-based. Therefore, we extend our evaluation to ask: can our method also outperform non-LLM metrics?

We use the official MTME toolkit[2] released by the WMT. To ensure fair comparison, we focus on reference-free metrics, since our method does not require gold references. Reference-based metrics introduce additional semantic input and belong to a fundamentally different evaluation paradigm. For completeness, we report reference-based results in Appendix A.

Together, the two experiments form a progressive evaluation: Experiment 1 compares our method to LLM-based direct scoring, while Experiment 2 benchmarks it against MTME's reference-free metrics, highlighting its broader advantages.

## 5 Results

Experiment 1 demonstrates that our method consistently outperforms the baseline proposed by Qian et al. (2024). Table 1 presents the results, where bolded values indicate cases in which our method outperforms the baseline. Three out of five models surpass the baseline across all five language pairs. The remaining two models outperform the baseline on three and two language pairs, respectively. These results suggest that strong performance on one language pair does not guarantee consistent effectiveness across others—model behavior can vary depending on the language. Overall, our method shows stronger and more consistent performance than the baseline within LLM-based evaluation settings.

Experiment 2 extends the comparison to reference-free MTME metrics. As shown in Table 2, bolded values mark cases where our method exceeds the average scores of MTME's [noref] systems. Among the four language pairs evaluated, our method achieves higher average correlations than all MTME baselines on three. Notably, in two of these three language pairs, our scores also surpass MTME's best-reported results. For the low-resource pair UK–EN, all of our models outperform all five MTME metrics—establishing new state-of-the-art results across the board.

This progression of results highlights not only the overall effectiveness of our method, but also its cross-linguistic generalizability across both high- and low-resource scenarios.

## 6 Conclusion

This work rethinks the role of large language models (LLMs) in machine translation quality estimation, shifting from direct scoring to a generation-based evaluation framework. We conduct the most extensive study to date, evaluating 8 LLMs and 8 language pairs through two experiments. Our method consistently outperforms LLM-based scoring approaches and surpasses non-LLM official MTME reference-free metrics, particularly in low-resource settings. These findings underscore the limitations of using LLMs as scorers and demonstrate the effectiveness of leveraging them as generators. We advocate for a hybrid evaluation paradigm that combines the fluency of generation with the semantic precision of embedding-based scoring.

---

[2] https://github.com/google-research/mt-metrics-eval

## Limitation

We acknowledge a key limitation of our study: all language pairs use English as the target language. This choice was made to reduce confounding variables and focus on the effects of the source language, especially given that prior studies have reached opposing conclusions about its role in LLM evaluation. While this design helped ensure internal consistency, we fully recognize that it limits the generalizability of our findings. We encourage future research to apply our method to a broader range of target languages to better assess its multilingual applicability.

## Ethics Statement

This study does not involve human subjects or sensitive content. All data used in our experiments are publicly available, including the WMT22 dataset from the WMT shared task, the dataset released in a prior EMNLP 2024 publication, and evaluation resources from the MTME toolkit. The language models employed are open-access and widely used in the research community. Our goal is to promote more effective and transparent approaches to machine translation evaluation. We believe this work offers constructive insights while adhering to ethical standards in data usage and model deployment.

# A Comparison with MTME Metrics

| | | UK-EN | | | CS-EN | | | RU-EN | | | DE-EN | | |
|---|---|---|---|---|---|---|---|---|---|---|---|---|---|
| | | ρ | r | τ | ρ | r | τ | ρ | r | τ | ρ | r | τ |
| ours | Gemma-7B | **0.025** | **0.016** | **0.017** | **0.041** | **0.030** | **0.028** | **0.000** | **0.005** | **0.003** | 0.011 | 0.006 | 0.004 |
| | Qwen3-8B | **0.019** | **0.006** | **0.013** | **0.047** | **0.038** | **0.031** | **0.009** | **0.004** | **0.006** | 0.009 | 0.011 | 0.006 |
| | Deepseek-R1 | **0.017** | **0.012** | **0.011** | **0.040** | **0.039** | **0.027** | **0.010** | **0.005** | **0.007** | 0.017 | **0.024** | 0.011 |
| | GPT-4-turbo | **0.016** | **0.010** | **0.011** | **0.033** | **0.038** | **0.022** | **0.006** | **0.004** | **0.004** | 0.012 | 0.017 | 0.008 |
| mtme [noref] | median | 0.005 | 0.004 | 0.003 | 0.026 | 0.024 | 0.018 | 0.002 | 0.007 | 0.001 | 0.024 | 0.020 | 0.016 |
| | mean | 0.004 | -0.001 | 0.002 | 0.024 | 0.027 | 0.017 | -0.001 | 0.003 | -0.001 | 0.022 | 0.021 | 0.015 |
| mtme [noref] | HWTSC-Teacher-Sim | 0.011 | 0.005 | 0.007 | 0.026 | 0.024 | 0.018 | 0.010 | 0.010 | 0.007 | 0.024 | 0.018 | 0.016 |
| | COMETKiwi | 0.005 | 0.005 | 0.004 | 0.041 | 0.044 | 0.028 | 0.002 | 0.008 | 0.001 | 0.017 | 0.020 | 0.011 |
| | UniTE-src | 0.005 | 0.004 | 0.003 | 0.038 | 0.039 | 0.026 | 0.002 | 0.007 | 0.001 | 0.027 | 0.030 | 0.018 |
| | REUSE | 0.000 | -0.008 | 0.000 | 0.002 | 0.005 | 0.002 | -0.011 | -0.010 | -0.007 | 0.014 | 0.012 | 0.009 |
| | COMET-QE | -0.003 | -0.010 | -0.002 | 0.015 | 0.022 | 0.010 | -0.007 | -0.002 | -0.005 | 0.030 | 0.024 | 0.020 |
| mtme | BLEU | 0.010 | 0.004 | 0.007 | 0.064 | 0.049 | 0.043 | 0.021 | 0.020 | 0.014 | 0.013 | 0.016 | 0.009 |
| | chrF | 0.004 | 0.000 | 0.003 | 0.062 | 0.048 | 0.042 | 0.022 | 0.019 | 0.015 | 0.025 | 0.028 | 0.017 |
| | BLEURT-20 | 0.003 | -0.001 | 0.002 | 0.053 | 0.046 | 0.036 | 0.021 | 0.024 | 0.014 | 0.026 | 0.031 | 0.018 |
| | COMET-20 | -0.003 | 0.002 | -0.002 | 0.050 | 0.044 | 0.034 | 0.020 | 0.025 | 0.014 | 0.027 | 0.025 | 0.018 |
| | YiSi-1 | 0.007 | 0.007 | 0.004 | 0.055 | 0.042 | 0.037 | 0.026 | 0.030 | 0.018 | 0.018 | 0.027 | 0.012 |
| | BERTScore | 0.004 | 0.005 | 0.003 | 0.058 | 0.046 | 0.039 | 0.028 | 0.033 | 0.019 | 0.016 | 0.026 | 0.011 |
| | COMET-22 | 0.002 | 0.005 | 0.002 | 0.046 | 0.050 | 0.031 | 0.019 | 0.031 | 0.013 | 0.028 | 0.038 | 0.019 |
| | MS-COMET-22 | -0.001 | 0.002 | 0.000 | 0.044 | 0.039 | 0.030 | 0.010 | 0.018 | 0.007 | 0.020 | 0.022 | 0.013 |
| | UniTE | 0.005 | 0.004 | 0.004 | 0.053 | 0.047 | 0.036 | 0.018 | 0.020 | 0.012 | 0.028 | 0.031 | 0.019 |
| | f200spBLEU | 0.009 | 0.003 | 0.006 | 0.064 | 0.049 | 0.043 | 0.027 | 0.024 | 0.018 | 0.014 | 0.020 | 0.010 |
| | metricx_xxl_MQM_2020 | 0.003 | 0.003 | 0.002 | 0.038 | 0.046 | 0.026 | 0.017 | 0.034 | 0.011 | 0.021 | 0.035 | 0.014 |